\documentclass[sigconf]{acmart}

\def\BibTeX{{\rm B\kern-.05em{\sc i\kern-.025em b}\kern-.08emT\kern-.1667em\lower.7ex\hbox{E}\kern-.125emX}}
    
\begin{document}

%
\title{Exploring Bias in GAN-based Data Augmentation 
for Small Samples}

%
\author{Mengxiao Hu}
\email{sa517122@mail.ustc.edu.cn}
\affiliation{
  University of Science and Technology of China
}
\author{Jinlong Li}
\email{jlli@ustc.edu.cn}
\affiliation{
  University of Science and Technology of China
}

%
\renewcommand{\shortauthors}{}

%
\begin{abstract}
For machine learning task, lacking sufficient samples mean trained model has low confidence to approach the ground truth function. Until recently, after the generative adversarial networks (GAN) had been proposed, we see the hope of small samples data augmentation (DA) with realistic fake data, and many works validated the viability of GAN-based DA. Although most of the works pointed out higher accuracy can be achieved using GAN-based DA, some researchers stressed that the fake data generated from GAN has inherent bias, and in this paper, we explored when the bias is so low that it cannot hurt the performance, we set experiments to depict the bias in different GAN-based DA setting, and from the results, we design a pipeline to inspect specific dataset is efficiently-augmentable with GAN-based DA or not. And finally, depending on our trial to reduce the bias, we proposed some advice to mitigate bias in GAN-based DA application.
\end{abstract}

%
%


%

%
\maketitle
\begin{figure}[t]
\centering
\includegraphics[width=70mm]{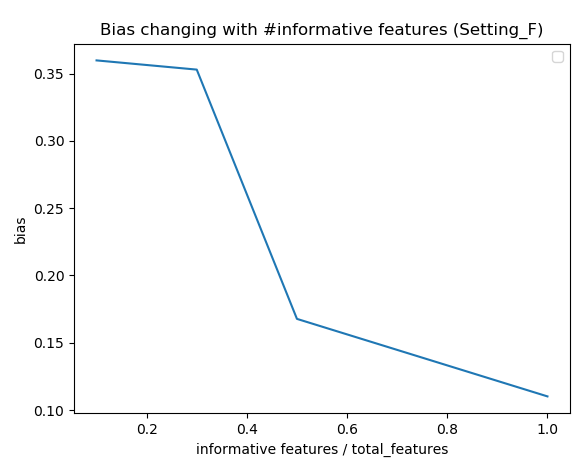}

\caption{Bias changing with the number of informative features. "Setting \textunderscore F" means the we assigned a consatnt number for each data's size. The size of each dataset is 10,500. We sample a small set of it for DA, using the same sampling size in figure 1 and 2. }\label{10}
\end{figure}
\begin{figure}[t]
\centering
\includegraphics[width=70mm]{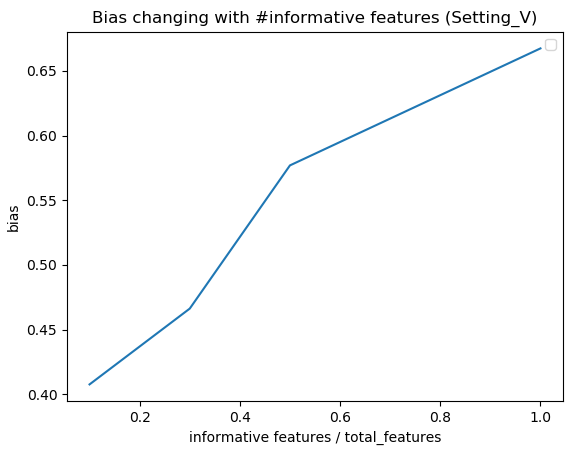}

\caption{ Bias changing with the number of informative features. "Setting \textunderscore V" means the we assigned a variant size(changing with the number of total features) for each dataset. In figure 1 and 2, we set the sizes of total features as 2,000 for all data. In this case, the variance setting seems meaningless, but we surprisingly found the changing trends are totally different with different data sizes. The size of each dataset is 25,510 here.}\label{11}
\end{figure}
\section{Introduction}

\subsection{Small Sample Size}

When the dimension of data exceeds the number of instances, we say the dataset has a small sample size.

If we have enough data, we can get a smaller sample data from it, but conversely, it is impossible to get more real extra data from a small dataset. The essence of machine learning algorithm is to discover a hypothesis near the ground truth function from little data, one of the advantages is that we need not to have higher cost to collect data to discretely depicts a function. But, training a learning model also needs enough data to prevent overfitting, so, when we have not sufficient data for a learning task, we can hardly use the small dataset. Small sample data is common in medical image diagnose, natural language processing and emotion classification. Researches in above area are important to us, therefore, we should make small sample data useful to machine learning algorithm.
\begin{figure}[t]
\centering
\includegraphics[width=80mm]{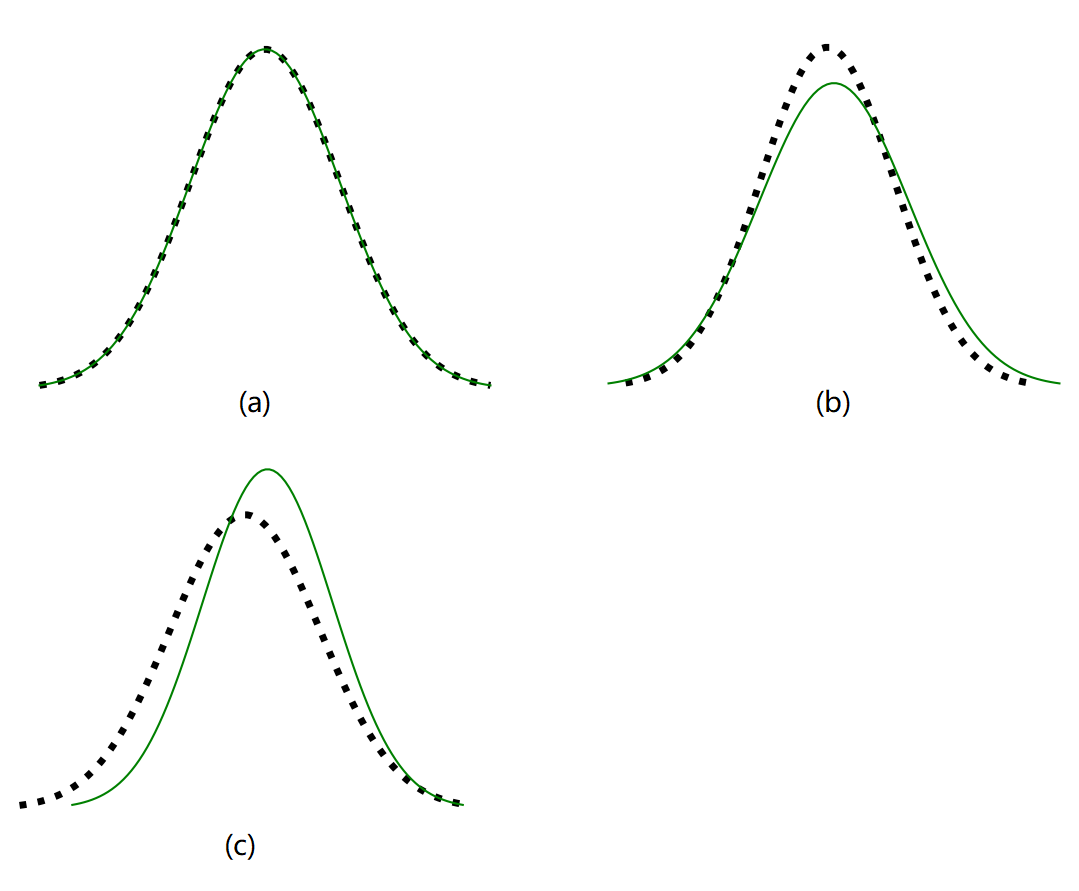}

\caption{(a) The theoretical generating distribution (green line) has the same shape as the original data distribution (dotted line), after it converged. (b) The real generating distribution has not the same shape, but has more diversity. (c) he real generating distribution has not the same shape, but has less diversity.}\label{1}
\end{figure}

\subsection{Data Augmentation}

There are several methods to augment a dataset, like rotation, flip and adding noise ~\cite{bjerrum2017smiles}, and advanced methods like conditional GAN that can generate winter landscape from summer landscape ~\cite{lee2018diverse}. The methods can be called data augmentation (DA), those DA methods have two advantages:
\begin{itemize}
\item They add instances which the model may encounter in the real world but have no similar ones in original dataset.
\item For deep learning algorithm, the methods add instances which are different but have same low-level features.
\end{itemize}
In this paper we only focus on one-layer machine learning problem, so, we cannot apply those DA methods for the second advantage. On the other hand, we want added instances from the same distribution of the original data, for example, if the original data lies on a manifold, we hope added instances lie on the same manifold. In other words, we want more diverse data from the same distribution. Because above DA methods generate data from other distribution, we cannot use them in our task for the first advantage.

\begin{figure}[t]
\centering
\includegraphics[width=80mm]{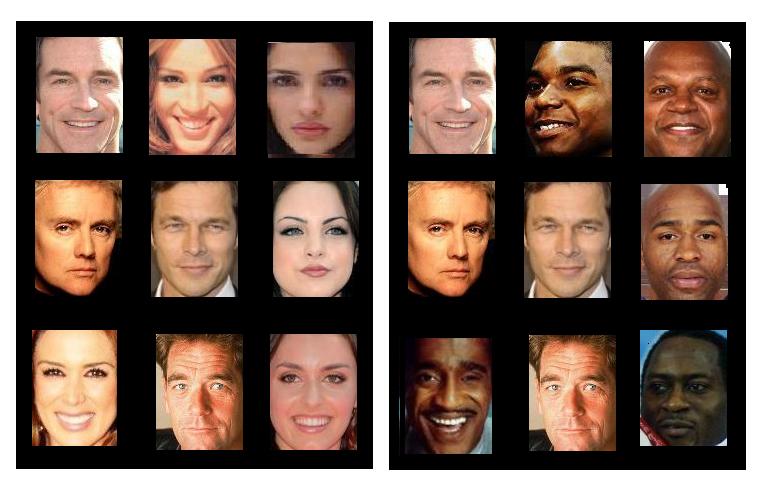}

\caption{Two face dataset, the left only white male and female, one has only male face white and black, the right dataset has higher variance than the left, because the difference of digital value between white and dark pixels is bigger the other color pairs, and the skin area is larger than other facial features'. But we cannot say an all-man dataset has more diversity, so, simple variance is not the diversity we want.}\label{2}
\end{figure}

\subsection{Generative adversarial networks}
 To generate data from the original distribution, generative adversarial networks was proposed ~\cite{goodfellow2014generative}. To ensure the generator produce high quality data, GAN has a discriminator to tell an instance is from the generator or the original dataset, and the generator should adjust generating distribution if generated instances are labeled as fake data, until the generating distribution is similar enough to the real distribution. This procedure is characterized as a two-player minimax game, and the value function is given as:
\begin{align}
\min _{G} \max _{D}V(D,G)= &\mathbb{E}_{\pmb{x}-p_{data}(\pmb{x})}[logD(\pmb{x})]\nonumber\\+&\mathbb{E}_{\pmb{z}-p_{\pmb{z}}(\pmb{z})}[log(1-D(G(\pmb{z})))]
\end{align}%

Theoretically, we can achieve its global optimum by gradient methods, and the final generating distribution is equal to original distribution as shown in figure 3-(a).

In real world, the value of loss functions in gradient methods cannot be down to zero. In our experiments, we found both discriminative loss and generating loss waver in a range after some iterations. When the loss values are not continuously decreasing, we accept the generating distribution is similar enough to the original one for our DA task. Notice that we want fake data is diverse, so, we checked generated data has bias or not. By observing the outputs digits images in different epoch during iteration, we found bias exits. We conclude that GAN can generate data from near real distribution, but cannot eliminate bias in data.  Goodfellow et al. ~\cite{goodfellow2014generative} pointed out "the discriminator must be synchronized well with the generator during training" to prevent the generator collapsing, when the generator is trained to have enough diverse fake data. We do not know whether the synchronization reduces the diversity or not. Although we can add a constraint 
\begin{align}
\max_{G}(A \,diversity \,\, measure) 
\end{align}
to formula (1), trying to constrain the bias, it is difficult to find a proper measurement for diversity. In this paper, we try to track the bias and to give practical advice on mitigating biases. We also found same GAN-based DA method performs differently on different small sample dataset. Therefore, we tried to design a pipeline for checking the suitability of applying GAN-based DA on a specific dataset.

\section{Related Work}

\subsection{Traditional Data Augmentation}

The term data augmentation is originated with Dyk et al. ~\cite{tanner1987calculation}'s Data Augmentation algorithm. But the data augmentation technique had been known as a tool in Markov chain Monte Carlo (MCMC) sampling methods ~\cite{dempster1977maximum}. The DA algorithm posits there is an augmented data to observed data mapping, which is a many-to-one mapping. It uses formula (3) to qualify this mapping
$\mathcal{M}:Y_{aug}\rightarrow Y_{obs}$:
\begin{align}
\int_{\mathcal{M}(Y_{aug})=Y_{obs}}p(Y_{aug}|\theta)\mu(dY_{aug}) \, = \, p(Y_{obs}|\theta)
\end{align}
In this formula, $Y_{aug}$ and $Y_{obs}$ refers to augmented and observed data, $\mu(\cdot)$ is a measure and $\theta \in \mathbb{R}^d$. Formula (3) tells us the marginal distribution of $Y_{obs}$ implied by $p(Y_{aug}|\theta)$ must be the original model $p(Y_{obs}|\theta)$, because the original model is not changed after the mapping being introduced. And we can set a initial value $\theta^{(0)}$, and then form a Markov chain $\{\theta^{(t)},Y_{aug}^{(t)}),t\geq1\}$ ~\cite{tanner1987calculation}. Zhang et al. ~\cite{van2001art} introduced a 'working parameter' into formula (3), and proposed marginal augmentation and conditional augmentation, then used a deterministic approximation method for selecting good DA mapping (or scheme). During the selection, to assess the performance of the proposed augmentation methods, they gave 2 criterions, both of them are related to the 'working parameters'. 

Because the field of machine learning was getting popular, and the generative model ~\cite{jebara2004generative} can generate data for augmentation, we presented related works in the field of generative models in the following sections.

\subsection{Generative Model}

From the statistical view, the relationship of augmented data and observed data is not treated as a many-to-one mapping, the relationship is depicted by the statistical modelling of the joint distribution on X and Y, here, X is an observable variable and Y is a target variable ~\cite{ng2002discriminative}.

Goodfellow \cite{goodfellow2016nips} pointed out that generative model can be learned via the principle of maximum likelihood, different models have different representations or constructions of the likelihood, and they can be classified into explicit density models and implicit density models. Pixel RNN ~\cite{oord2016pixel}, Variational autoencoder ~\cite{kingma2013auto} and Boltzmann machine ~\cite{fahlman1983massively} are explicit density models, GAN ~\cite{goodfellow2014generative} and Generative Stochastic Networks (GSN) ~\cite{alain2016gsns} are implicit density models. Goodfellow points out GAN has not disadvantage of other models, because it searches a Nash equilibrium of a game rather than optimizing an objective function. Goodfellow also points find a Nash equilibrium of a game is more difficult.

We believe our work can reveal the disadvantage of GAN to some extent, and can remind people that sampling real-world data cannot totally be substituted by generated data of GAN, even though GAN-based DA can sometimes boost performance of a model ~\cite{perez2017effectiveness}.

For the clarity of the terminology, the 'augmented data', 'fake data' and 'generated data' share the same meaning, the 'observable data' and 'real data' share the same meaning, but we use different terms in different contexts in this paper.

\subsection{GAN-based Data Augmentation}

There are many work used GAN-based DA for small sample data. Ramponi et al.~\cite{ramponi2018t} used it to on time series data. Hu et al.~\cite{hu2018prostategan} used it on cancer images. Chang et al.~\cite{chang2018code} used it on NLP data. Liu et al.~\cite{liu2018pixel} used it on semantic image data. Bowles et al.~\cite{bowles2018gansfer} also used it on medical image data. Zhang et al.~\cite{zhang2018dada} modified vanilla GAN and used DA methods on several classification dataset. Lim et al.~\cite{lim2018doping} used it on anomaly detection data. They showed GAN-based DA is an efficient way to improve model performance, but they did not answer that does more generated data, or, data from longer trained generator improves performances further? 

Jain et al.~\cite{jain2018imagining} pointed out that data biases are increasing with longer training generator in DCGAN-based DA.   But, it should provide a universal bias measurement, and should valid the discover on more datasets and more GAN's variants.

\subsection{Debiasing Algorithm}

Recently, Amini et al. ~\cite{amini2019debias} propose an algorithm, debiasing VAE (DB-VAE), on mitigating bias, but it needs a large dataset to learn the latent structure of that data. In this paper, we explored the bias of data that generated by GAN. We believe our work is more fundamental, because even we can get a large dataset by GAN-based DA, and then implement DB-VAE on it, we cannot ensure the generated data is low-bias enough to represent the structure of the original small dataset. In other words, if the bias of generated data is too high, using DB-VAE is not efficient, for example, if we must sample millions of faces from a generated distribution to get a black face, we can reweight it use DB-VAE, but we cannot afford the computation. 

\begin{figure}[t]
\centering
\includegraphics[width=80mm]{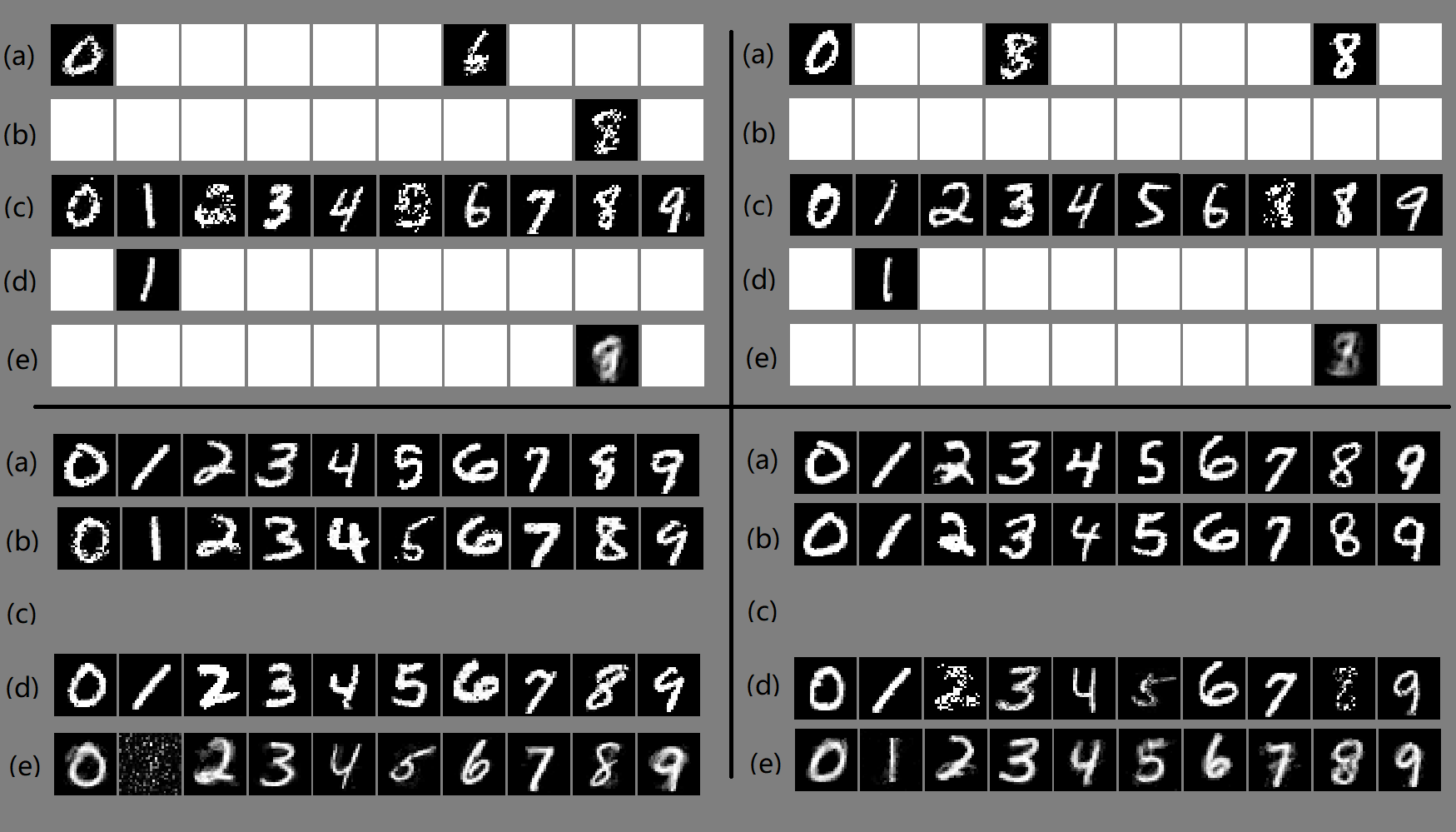}

\caption{The upper figures are results from using mixed data, the lower figures are results from recombination of one-group generating data. From (a) to (e) are the results from softmax GAN, info GAN, conditional GAN, boundary seeking GAN and boundary equilibrium GAN. The right figures are the results after 50,000 iteration training, the upper/lower left figure is the result after 20,000/10,000 iteration training, the white blocks means that digit is not in 160 generating instances.}\label{3}
\end{figure}
\begin{figure*}[t]
\centering
\includegraphics[width=160mm]{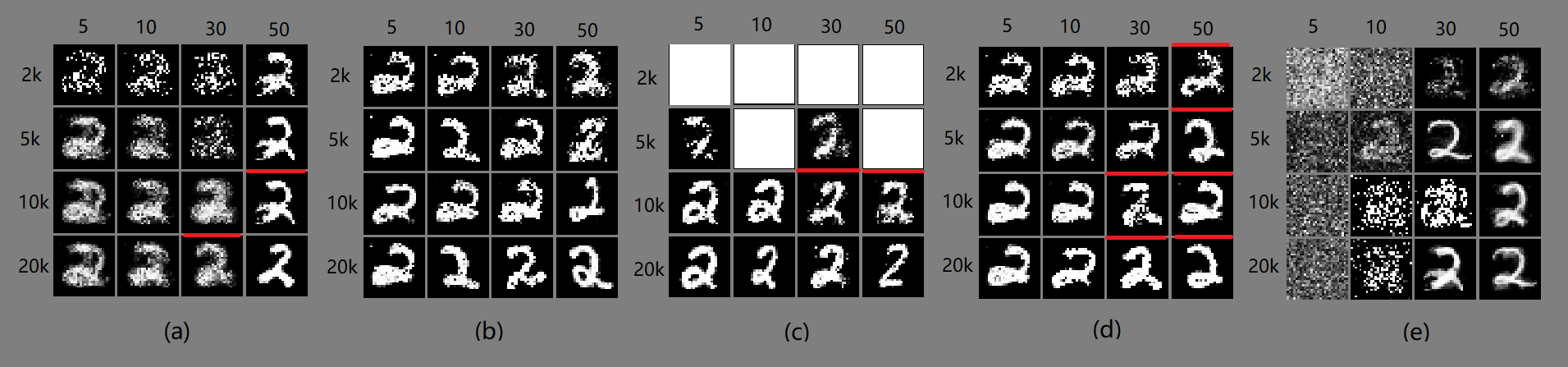}

\caption{Feed one class data to GAN, results from different sample sizes (5, 10, 30, 50) and training iterations (2,000, 5,000, 10,000, 20,000), and (a) to (e) refers the same variants as in figure 5. For each combination of parameters, we generated 16 samples and randomly picked one to a small block, the one has a red bar above it means its 16-images parent set has observable diversity.}\label{4}
\end{figure*}

\subsection{Contributions}
In the task of landmarking and simple classification on chosen small samples dataset, we showed:
\begin{itemize}
\item Different performances of 5 GAN variants as DA methods on different datasets.
\item A different way to sample instances from fake data, and its effects on DA.
\item When we using softmax GAN to augment data, the bias is monotone correlated with the number of the informative features of the data.
\item A pipeline was designed to check if simple GAN-based DA is suitable for a specific dataset or not.
\item Several pieces of advice were suggested to mitigate bias in GAN-based DA application.
\end{itemize}

\section{Experiments and Results}
\subsection{Data Bias Measurements}
In our paper, data bias and data diversity are inverse correlated concepts. To the best of our knowledge, no theoretical definition of data diversity has been proposed. Even though the bias of a classifier was defined ~\cite{amini2019debias}, we believe the bias is inherited in the data. We cannot simply use statistical variance to measure data diversity, shown in figure 4. If we only use simple data variance to measure data diversity, some features may be overlooked, like the gender features. So, they should have more weight in measuring diversity, but gender cannot be easily decided by facial attributes ~\cite{rozsa2016facial}. 

For labeled data, diversity can also be characterized by the density of each label group.  Apart from few GAN variants like conditional GAN, no variant can generate labeled fake data, so, if we want to compute the density, generated data should be labeled first. We can use clustering algorithm to group unlabeled data, and compute the density from number of instances and the diameters of the cluster. But, if data cannot be easily clustered in fixed-axes ellipses, we can hardly use CURE ~\cite{guha1998cure} to have irregular shaped clusters, because calculating clusters' area is difficult. And if we use BFR ~\cite{bradley1998scaling} or k-means algorithm ~\cite{forgy1965cluster}, computing density of a group from several sub-clusters is not easy, because inter-cluster distance must be considered.

Considering the data is used to train a classification model in this paper, we found a modeling error ~\cite{bishop_2016} that can be used to measure data biases. Because it does not directly use original data to compute biases, it has no problems above.

\textbf{Supposition 1.} \textit{For curve fitting problem, if training data and test data from the same distribution, and if training data are evenly distributed, the more training data, the overfitting is less.}

In classification task, overfitting can be defined as prediction accuracy difference between training data and test data.
\begin{align}
Overfitting={Accuracy_{\,training}}-{Accuracy_{\,test}}
\end{align}

In our experiments, we only use definition (4) to measure data biases.
From supposition 1., we believe overfitting is a proper  measure for data biases, in other words, it can measure data diversity. But we cannot apply it to formula (2), because prediction accuracy needs time-consuming model training, and we can hardly use it in gradient method.

\subsection{Experiments on GAN's Variants}
Jain et al.~\cite{jain2018imagining} demonstrate DCGAN-based DA can perpetuate data biases, and they declare GAN-based DA perpetuates data biases. But DCGAN is only one of the GAN's variants, we cannot conclude all GAN-based DA perpetuate data biases. To verify data biases are perpetuated by other GAN's variants-based DA, we randomly choose five GAN's variants in this section, they are softmax GAN ~\cite{lin2017softmax}, info GAN ~\cite{chen2016infogan}, conditional GAN ~\cite{mirza2014conditional}, boundary seeking GAN ~\cite{hjelm2017boundary} and boundary equilibrium GAN ~\cite{berthelot2017began}.
\subsubsection{Least Input Size for GAN}
Because GAN's generator is to learn original data's distribution, and too few instances cannot discretely represent the distribution well, we must find the least input size for GAN for the following experiments. 

At the beginning, we used mixed classes data to augment, but most GAN's variants can only generate few classes, in other words, the biases are too high to evenly sample different classes, as shown in figure(3). Apart from conditional GAN, the other variants cannot generate images from given labels, so, we group the data by the labels, then feed different groups respectively to the variants. And we use this training method for other four variants in all following experiments.

To quickly judge the quality of generated data, we used MNIST dataset for the experiment, and focus on one digit's images at first. Because we do not know how many epochs are enough for training iteration, we set different iterations in the experiment.

From the figure 6, we believe 50 is big enough as the size of one group, and 10,000 is big enough as the iteration. In the following experiments, the least size refers to 50, and the least iteration refers to 10,000.
\subsubsection{Tracking Bias Changes}
When the training is end, we used the generator to sample fake data from fake distribution. The figure 7 shows the biases changing with different iteration ends and sample sizes.

We cannot observe positive or negative correlations between biases and iterations. It is possible that, after the generating distribution converge to some point, ,when the training is continuing, it wavers around the point, and the bias wavers in some interval along with the distribution.

In the figure 9, the biases from different variants are all increasing with the sample size, and slowly converges to some values, which means the generators maybe evenly sample data from fake distributions.
\begin{figure}[t]
\centering
\includegraphics[width=80mm]{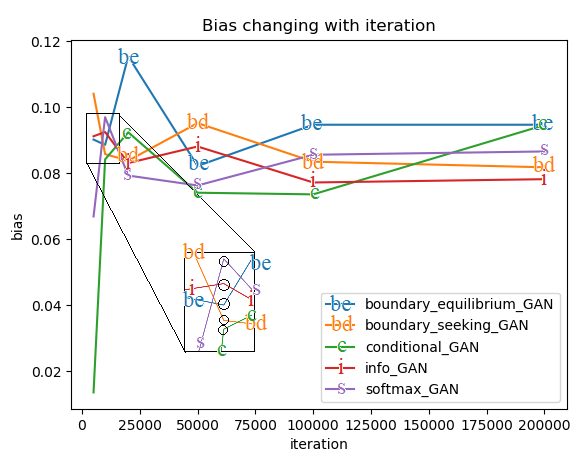}

\caption{Bias changing with iteration, (iteration: 5,000, 10,000, 20,000, 50,000, 100,000, 200,000). When iteration is 10,000, the biases of purple (0.0970), red (0.0925) and green (0.0842) are on the peaks, the biases of blue (0.0886) and yellow (0.0858) are in the valleys.}\label{5}
\end{figure}
\subsubsection{Sampling Fake Data}
figure 7 shows that, longer training generators do not generate lower/higher biases data. So, it is difficult to decide an iteration end, when we use one-shot sampling to get all augmentation data. 

However, if we sample a batch of data during each epoch, we will not fill up with the augmentation dataset with only highest-bias data. But it is not practical in our experiments, because the augmentation data size will be very large if we sample data during each epoch. So, we sample a batch after every n iterations. On the one hand, during early epoch, the generating distribution has not converged, we set a sampling start epoch in our experiments, on the other hand, because we want to compare mixed sampling results with one-shot sampling ones, we should not sample data 
\begin{figure}[H]
\centering
\includegraphics[width=80mm]{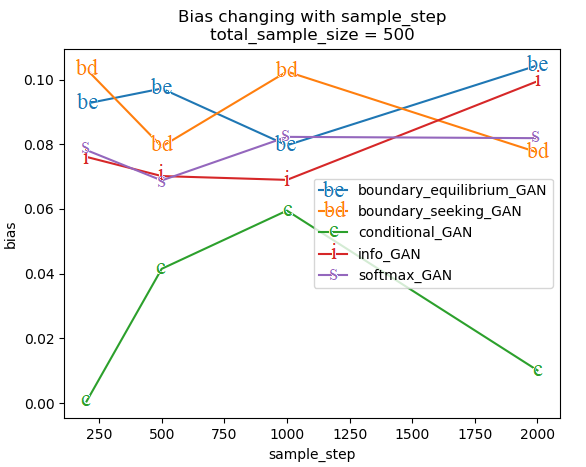}

\caption{Using mixed sampling, iteration ends at 10,000 epoch, bias changing with sample step, (sample step: 200, 500, 1,000, 2,000). The biases of purple (0.0782), red (0.0761) and green (0.0005) are lower than the peaks', blue (0.0926) and yellow (0.1032) are higher that the valleys', mixed sampling seems to 'average' the biases at different iteration.}\label{7}
\end{figure}
wholly at epochs which are larger/lower than the least iteration. Therefore, we choose 5,000 as the sampling start epoch and 15,000 as the sampling end point. 

\begin{figure}[t]
\centering
\includegraphics[width=80mm]{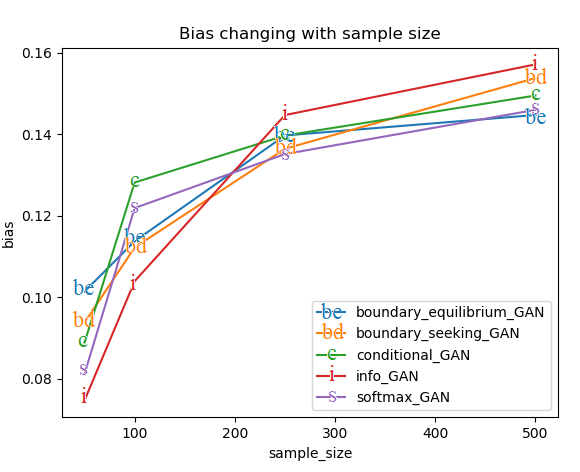}

\caption{Biases changing with sample size, here, sample size is one class sample size (50, 100, 200, 500), because MNIST has 10 classes, the total sample size is from 500 to 5,000.}\label{6}
\end{figure}

To keep the augmentation data sizes(called total sample size in figure 8) same as previous experiments', we set the batch size as a variance.

In figure 7 and figure 8 we set the one class sample size 50, we found mixed sampling have 'averaging biases' ability.  More specifically, in figure 7, we see that when iteration is 10,000, the purple, red and green points are in the peaks, the average of biases in [5000, 15,000] should be lower than the valleys', and for blue and yellow points, the average of biases in [5000, 15,000] should be higher than the valleys' value'. 

In fact, in figure 8, when sample step is 200, all points are in predicted positions, we see the biases of purple, red and green are smaller than the peaks' value in figure 7, and the biases of blue and yellow points are higher than the valleys' value. However, we cannot explain why the biases are changing with the sample step.

\begin{figure}[t]
\centering
\includegraphics[width=80mm]{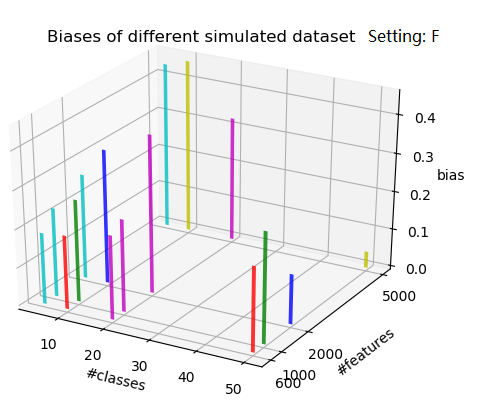}

\caption{When generating simulated classification data, we set the number of instances fixed, a constant 500. The biases are increasing with the number of features, if per class has enough instances(samples). The bars' colors are for easy read.}\label{8}
\end{figure}
\subsection{Experiments on Different Datasets}
To verify GAN-based DA also perpetuate data biases on other dataset, in this section, we applied softmax GAN-based DA on simulated data set and four GAN's variants-based DA on four real-world dataset.

\subsubsection{Simulated Data}
We used the generating function in scikit-learn package ~\cite{scikit-learn} to make data. The function has 15 parameters for generating data, we picked several important parameters as controlled variance. Because we are focusing on data with small sample size, we picked the number of features. We did not pick the number of samples as controlled variance, because we think the ratio of samples and features is more important, considering the definition of small sample data. Because we generated one class data and then mix them as the augmentation data, the bias of the augmentation dataset is the 'combination' of different classes' biases, so we picked the number of classes as the second controlled variance. The last controlled variance we picked is the number of informative features, like the facial features in face dataset. Intuitively, we say two faces are similar when they have similar facial features, if a face dataset has low diversity, many faces in it are similar. 

We designed two group of experiments, each test the three controlled variances. One of them has changed ratio of samples and features, the other has fixed ratio. Experiments results are shown in figure 10, 9, 1 and 2.

From figure 11, we cannot find a regular relationship between biases and the number of classes or the number of features, when the ratio of samples and features is not changing.

In figure 10, the ratio of samples and features is increasing with the number of features. And we found when the number of classes is less than 20, the biases are increasing with the features. To illustrate the strange trend at the number of classes 50, we notice that the augmentation data size is constant 500, so each class group has only 10 instances, and it may be under the specific least input size for softmax GAN-based DA. 

In the experiments of bias changing with the number of informative features, we set the dimension(total features) of all instances as 2,000. In this case, although 'Setting\textunderscore V' seems meaningless, but we surprisingly found the changing trends are totally different with different data sizes. Unluckily, we cannot explain how they are formed. 
\begin{figure}[t]
\centering
\includegraphics[width=80mm]{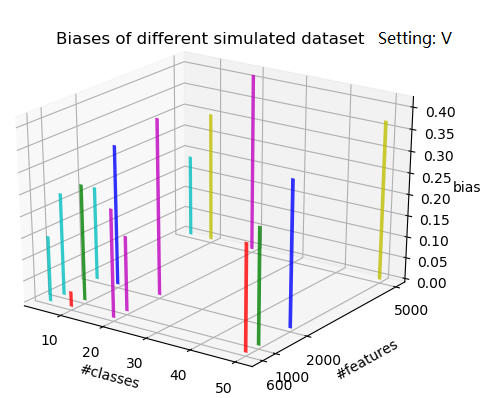}

\caption{When generating simulated data, we set the number of instances(samples) as a variance, which equals to (\#features)·(500/784).}\label{9}
\end{figure}
\subsubsection{Real-World Data}
We applied four GAN's variants DA respectively on Parkinson's Disease Classification Data Set ~\cite{sakar2019comparative}, SCADI Data Set ~\cite{zarchi2018scadi}, Amazon Commerce reviews set Data Set ~\cite{liu2011application} and a subset of the CIFAR-100 dataset ~\cite{krizhevsky2009learning}. 

Apart from the CIFAR-100 dataset, we shuffled each dataset, and then split each to training set and test set, the ratio of two sets' size is 1:1. 

For the CIFAR-100 dataset, we extract all instances under superclass (also called coarse class) 'aquatic mammals' as the total dataset, and keep their fine classes as their labels. 
Then we treat it like other datasets.

We used one-shot sampling to generate fake data. For all dataset, we setting iteration end after 20,000 epoch, and sample 50 fake data for each class.

Using this setting, we surprisingly find that (shown in figure 12), when softmax GAN, conditional GAN and boundary seeking GAN applied to the SCADI dataset, no positive bias was measured out. 
Because SCADI dataset is very sparse, and is not a image dataset, we cannot observe the augmentation data has bias or not.

\begin{figure}[t]
\centering
\includegraphics[width=80mm]{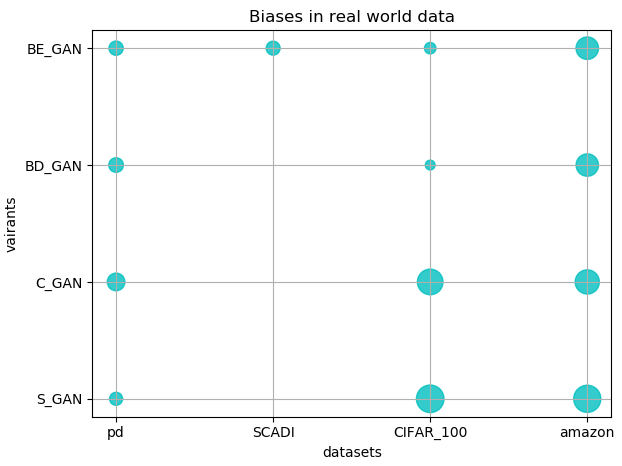}

\caption{Not all GAN's variants-based DA exaggerate data biases on all data.}\label{12}
\end{figure}

\subsection{A Pipeline to Check Augmentable Dataset}
From above results, we designed a pipeline (shown as 13) to estimate the viability of GAN-based DA on specific dataset. Suppose that the data is for classification task.
\subsection{Advice to Mitigate Bias in GAN-based DA}
From above results, we conclude some advice that might help you mitigate the bias when you use GAN-based DA.
\begin{itemize}
\item When deciding the iteration end, at the beginning, use mixed sampling to estimate the average biases in an iteration interval, then try a one-shot sampling, if the new bias is larger than the average, pick another iteration end for the next one-shot sampling, until the latest bias is below the average.
\item Try different GAN's variants at once, because from figure 12, we know for some dataset, some GAN's variant-based DA does not perpetuate bias.
\item Sample little data from generating distribution, if you find a larger sampling data has a larger bias.

Although according to supposition 1., we want a larger augmentation dataset, but in figure 9, we see when the generating distribution has shape as figure 3(c), the larger augmentation data has larger bias.
\item Try use GAN's variant which can output given label data, like conditional GAN. 

Because, the mixed-classes data has more information than one-class data. And when recombined one-class generating data, you must design the size of each class fake data, to keep the same classes distribution as the original dataset.
\section{Conclusion and Future Work}
In this paper, we are focusing on GAN-based data augmentation on small sample data, in classification task. We conducted several experiments on biases of augmentation datasets. At the beginning, we sought a measurement of data bias, and found the classification accuracy can be used, as shown in formula (4). Then, in the experiments on: biases changing with iteration of GAN's training, biases changing with sample size of fake data. We found that after the generating distribution converged, longer training of generator does not always reduce biases or increase biases, and when the generating distribution has bias, the larger augmentation dataset has larger biases. Depending on the findings, we tried sampling data during different epochs, and found that can average biases of one-shot sampling. In the experiments on simulated data, we found biases are high when the original training data has a high ratio of samples and features, and if two dataset has the same ratio of samples and features, the one has more informative features has lower biases of its augmentation data. And in the experiments on real-world data, we found some variants-based DA does not perpetuate bias in some dataset. Finally, according to the results, we designed a pipeline to estimate the viability of GAN-based DA on a dataset, and gave some advice on mitigating bias when using GAN-based DA. 

However, some work we cannot do in this paper. One part of the reason is that, almost no measurement of bias has been proposed and approved, so, it is desirable to design more bias measurements. Another part of the reason is that, there are many GAN's variants has been proposed and being proposed, different synchronization of the discriminator and the generator may have different effects of 'reducing diversity', as I mentioned in the introduction section., therefore, we should test more GAN's variants. If there is one of the variants immune to bias, we may discover a universal way to eliminate or mitigate bias in other GAN-based DA methods.
\end{itemize}
\section{Acknowledgement}
Anonymous for Reviewing.

\begin{figure}[tp]
\centering
\includegraphics[width=60mm]{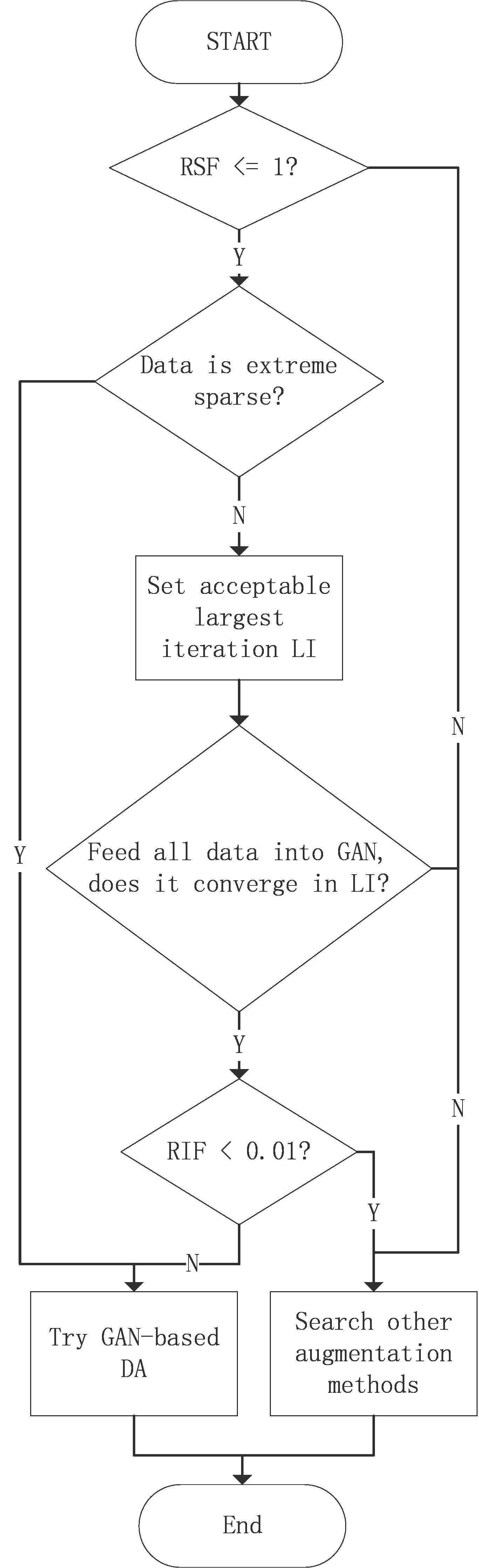}

\caption{The pipeline to estimate a dataset augmentable or not. RSF refers to the ratio of samples and features, RIF refers to the ratio of informative features and total features}\label{13}
\end{figure}

\bibliographystyle{named}
\bibliography{mybib}{}

\end{document}